\def\year{2021}\relax
\documentclass[letterpaper]{article} 
\usepackage{aaai21}  
\usepackage{times}  
\usepackage{helvet} 
\usepackage{courier}  
\usepackage[hyphens]{url}  
\usepackage{graphicx} 
\usepackage{natbib}  
\usepackage{caption} 
\usepackage{amsmath}
\usepackage{multirow}
\usepackage{colortbl}
\usepackage{amsthm}
\usepackage{booktabs}
\usepackage{amssymb}
\usepackage[switch]{lineno}

\definecolor{mygray}{gray}{.9}
\newcommand{\tabincell}[2]{\begin{tabular}{@{}#1@{}}#2\end{tabular}}

\title{Topic-Oriented Spoken Dialogue Summarization for Customer Service \\ with Saliency-Aware Topic Modeling}
\author{
    Yicheng Zou,\textsuperscript{\rm 1}
    Lujun Zhao,\textsuperscript{\rm 2}
    Yangyang Kang,\textsuperscript{\rm 2}
    Jun Lin,\textsuperscript{\rm 2}
    Minlong Peng,\textsuperscript{\rm 1}
    Zhuoren Jiang,\textsuperscript{\rm 3}
    \\
    Changlong Sun,\textsuperscript{\rm 3,2}
    Qi Zhang,\textsuperscript{\rm 1}
    Xuanjing Huang,\textsuperscript{\rm 1}
    Xiaozhong Liu\textsuperscript{\rm 4}
    \\
}
\affiliations{
    \textsuperscript{\rm 1}School of Computer Science, Fudan University, Shanghai, China\\
    \textsuperscript{\rm 2}Alibaba Group, China\\
    \textsuperscript{\rm 3}Zhejiang University, Hangzhou, China\\
    \textsuperscript{\rm 4}Indiana University Bloomington, Bloomington, United States\\

    \{yczou18, mlpeng16, qz, xjhuang\}@fudan.edu.cn, \{lujun.zlj, yangyang.kangyy, linjun.lj\}@alibaba-inc.com, \\
    jiangzhuoren@zju.edu.cn, changlong.scl@taobao.com, liu237@indiana.edu

}

\begin{document}

\maketitle

\begin{abstract}

In a customer service system, dialogue summarization can boost service efficiency by automatically creating summaries for long spoken dialogues in which {\em customers} and {\em agents} try to address issues about specific topics. In this work, we focus on topic-oriented dialogue summarization, which generates highly abstractive summaries that preserve the main ideas from dialogues. In spoken dialogues, abundant dialogue noise and common semantics could obscure the underlying informative content, making the general topic modeling approaches difficult to apply. In addition, for customer service, role-specific information matters and is an indispensable part of a summary. To effectively perform topic modeling on dialogues and capture multi-role information, in this work we propose a novel topic-augmented two-stage dialogue summarizer (TDS) jointly with a saliency-aware neural topic model (SATM) for topic-oriented summarization of customer service dialogues. Comprehensive studies on a real-world Chinese customer service dataset demonstrated the superiority of our method against several strong baselines.
\end{abstract}

\section{Introduction}
In an active customer service system, massive dialogues conveying important information between {\em customers} and {\em agents} are generated in real time. With this background, how to efficiently consume dialogue information becomes a non-trivial issue. Dialogue summarization is a task that aims to condense dialogues while retaining the salient information \cite{rambow2004summarizing,pan2018dial2desc,shang2018unsupervised,liu2019automatic}, which can boost service efficiency by 
automatically creating concise summaries to avoid time-consuming dialogue reading and comprehension.
\begin{figure}[t]
\centering
  \includegraphics[width=2.8in]{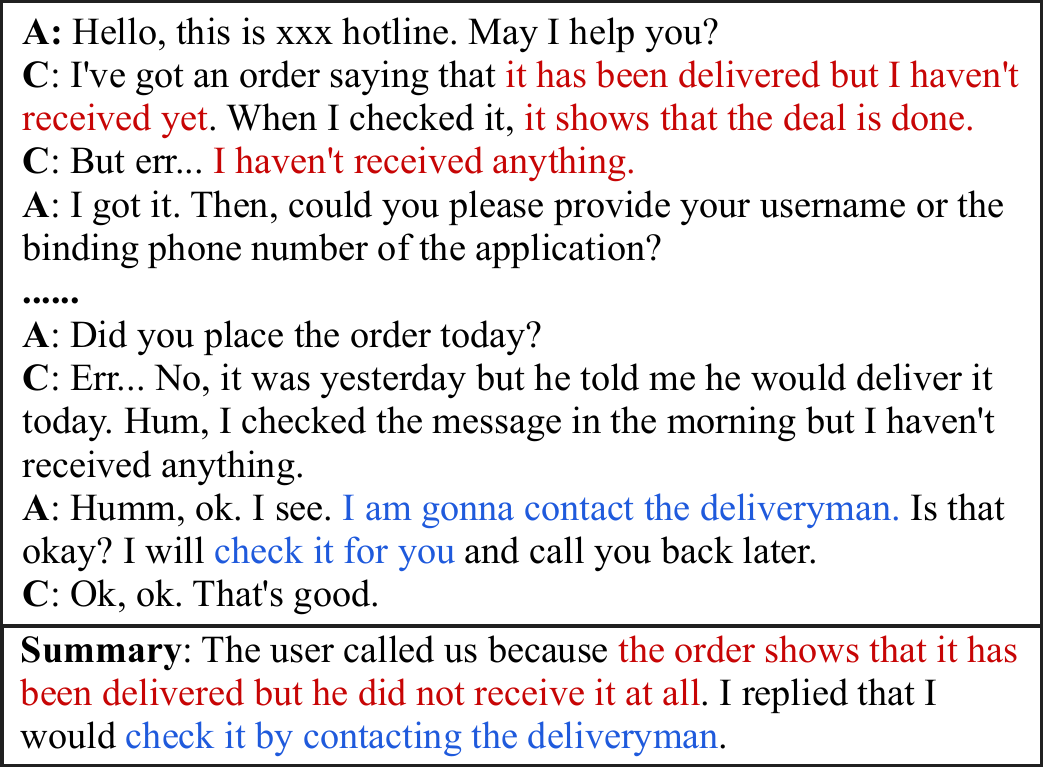}
  \caption{A customer service dialogue and its reference summary. {\bf C} denotes the customer and {\bf A} denotes the agent. The summary contains the customer's problem and the agent's solution, which are highlighted in red and blue, respectively.} \label{fig:intro}
\end{figure}

Most existing works for dialogue summarization have mainly focused on long and intricate spoken dialogues, like meetings and court debates, which are usually summarized by stringing all dialogue points to maintain an integral conversation flow \cite{gillick2009global,shang2018unsupervised,duan2019legal}. Nevertheless, in the customer service scenario, dialogue speakers commonly have strong and clear motivations and aim to address issues about specific topics \cite{wang2020sentiment}. To better understand customers' and agents' intentions, in this work we focus on the topic-oriented dialogue summarization, which aims to extract semantically consistent topics and generate highly abstractive summaries to maintain the main ideas from dialogues.

\begin{figure*}[t!]
\centering
  \includegraphics[width=6.0in]{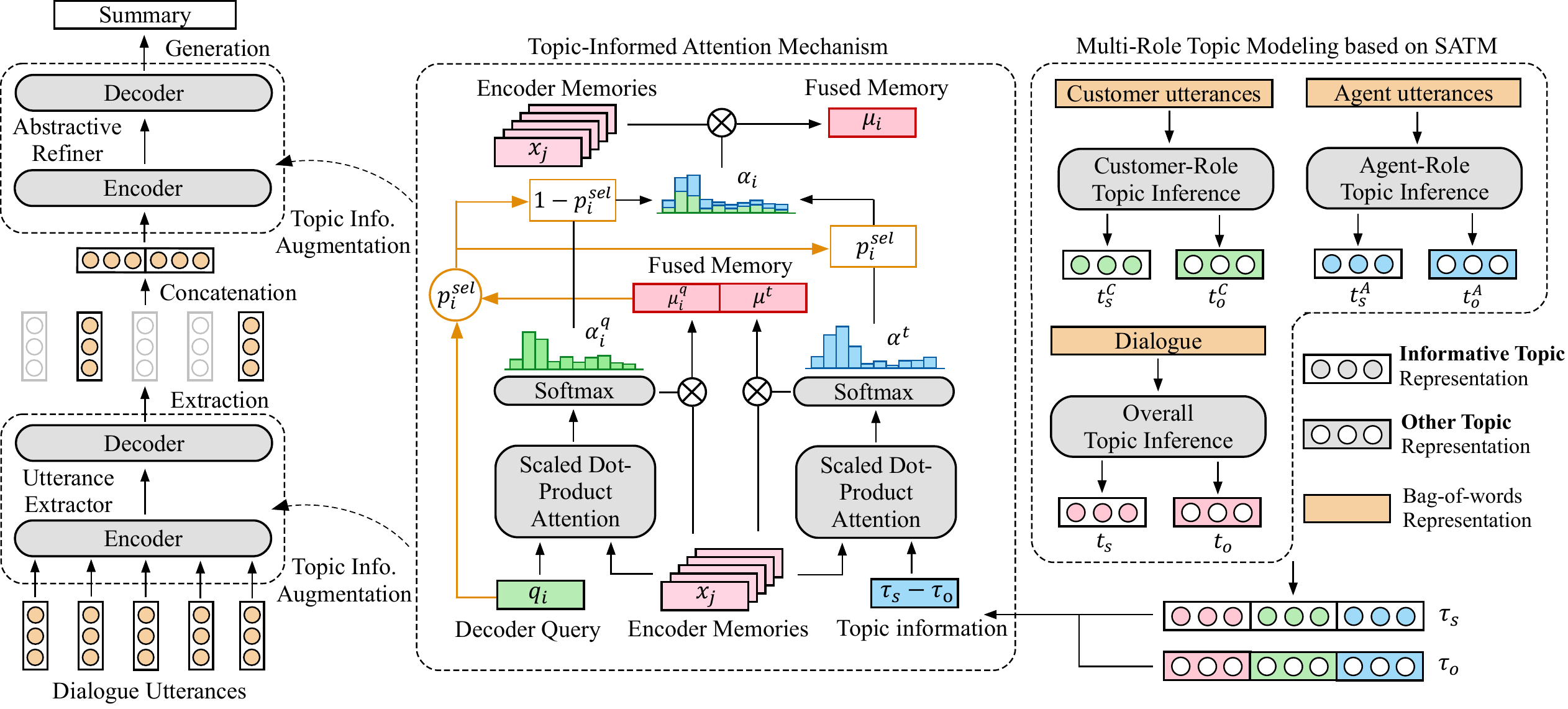}
  \caption{Overview of our proposed TDS with multi-role topic modeling based on SATM.} \label{fig:model}
\end{figure*}
Recently, dozens of topic-aware models have been introduced to assist with document summarization tasks \cite{wang2018reinforced,narayan2018don,fu2020document}. However, rather than well-formed sentences found in conventional documents, spoken dialogues are often composed of {\em utterances}. Salient information is diluted across these utterances and is accompanied by common semantics. Additionally, noise abounds in the form of unrelated chit-chats and transcription errors \cite{tixier2017combining}. Such common or noisy words, e.g., {\em please}, {\em thanks}, and {\em humm}, usually have a high frequency and co-occur with other informative words. As a result, the general topic-based approach can hardly distinguish the mixture of useful and useless content statistically, leading to inaccurate estimations of topic distribution
\cite{li2018filtering,li2019dirichlet}. Besides, in a customer service dialogue, the participating roles are stable: a customer tends to raise a problem and an agent needs to provide solutions. Figure \ref{fig:intro} shows a real-world customer service dialogue along with a summary that includes critical information from the two speakers. Hence, the model is also expected to capture role information to assist with saliency estimation.

In this work, we propose a novel two-stage neural model jointly with an enhanced topic modeling approach for spoken dialogue summarization. {\bf First}, to better distinguish the underlying informative content from abundant common semantics and dialogue noise, we introduce a saliency-aware topic model (SATM), where topics are split into two groups: {\em informative topics} and {\em other topics}. In the generative process of topic modeling, we constrain each salient word that corresponds to the gold summary to be generated from {\em informative topics}, while other words in the dialogue (including noisy and common words) are generated only from {\em other topics}. Through this training process, SATM can associate each word in a dialogue with either saliency (informative topics) or not salient (other topics). {\bf Second}, to capture role information and extract semantic topics from dialogues, we employ SATM to perform multi-role topic modeling on customer utterances, agent utterances, and overall dialogues separately. Then, a topic-augmented two-stage dialogue summarizer (TDS) is designed, which consists of an utterance extractor and an abstractive refiner. It can pick out topic-relevant salient information on both the utterance level and word level via a topic-informed attention mechanism. 

Furthermore, due to the lack of suitable public benchmarks, we collected a real-world customer service dialogue dataset with highly abstractive summaries. Experimental results on the proposed dataset showed that our model outperforms a series of strong baselines under various metrics. Codes, datasets, and supplementary can be found at Github\footnote{https://github.com/RowitZou/topic-dialog-summ}.

In summary, our contributions are as follows: 1) We introduce a novel topic model that can perceive underlying informative content in dialogues by directly learning word-saliency correspondences. 2) Based on multi-role topic modeling, we propose a topic-augmented two-stage model with a topic-informed attention mechanism to perform saliency estimation and summarize customer service dialogues.
3) Experimental results on the collected dataset demonstrate the effectiveness of our method in different aspects.

\section{Method}
In this section, we will detail the saliency-aware topic model (SATM) and the topic-augmented two-stage dialogue summarizer (TDS). The SATM infers multi-role topic representations based on {\em informative topics} and {\em other topics}. Then topic information is incorporated into the extractor and the refiner of TDS via a topic-informed attention mechanism. The overall architecture of our model is shown in Figure \ref{fig:model}.

\subsection{Saliency-Aware Neural Topic Model}

\begin{figure}[t!]
\centering
  \includegraphics[width=3.0in]{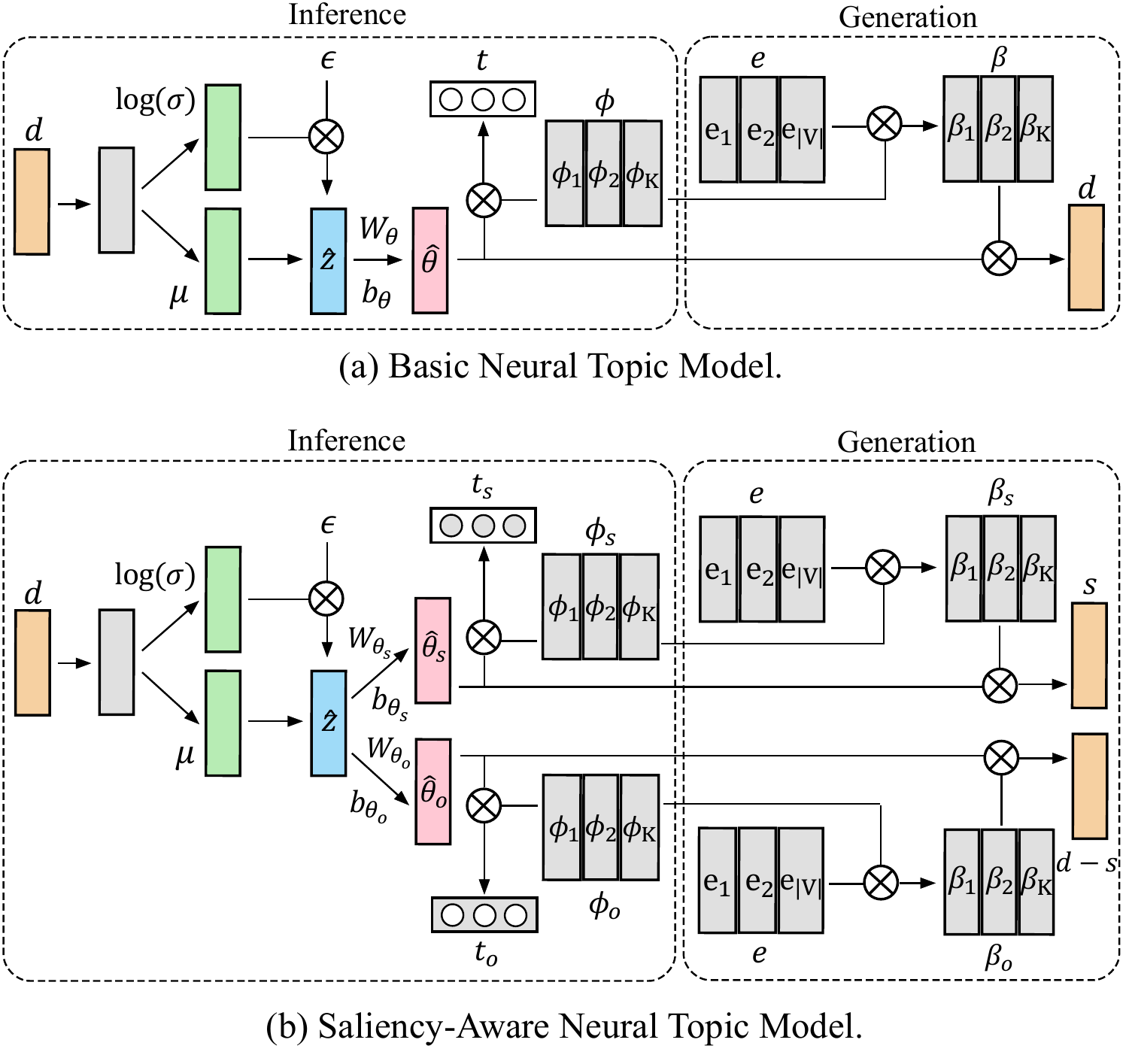}
  \caption{Comparison of NTM and SATM.} \label{fig:topic}
\end{figure}

Our proposed SATM is based on the Neural Topic Model (NTM) with variational inference \cite{miao2017discovering}, which infers the topic distribution $\theta$ from each dialogue $d$ by a neural network. We extend NTM with a new generative strategy to learn the word-saliency correspondences. The architecture of SATM compared with NTM is shown in Figure \ref{fig:topic}.

{\bf Basic NTM with Variational Inference.} Formally, given the bag-of-words representation of a dialogue $d\in\mathbb{R}^{|V|}$ with stop words removed, we build an inference network $q(\theta|d)$ to approximate the posterior $p(\theta|d)$, where $V$ is the vocabulary. $q(\theta|d)$ is composed of a function $\theta=f(z)$ conditioned on a diagonal Gaussian distribution $z\sim \mathcal{N}(\mu(d),\sigma^2(d))$, where $\mu(d)$ and $\sigma(d)$ are neural networks. In practice, we can sample $\hat{z}$ using a re-parameterization trick \cite{kingma2014auto} by $\hat{z} = \mu(d) + \epsilon \cdot \sigma(d)$, where $\epsilon$ is sampled from $\mathcal{N}(0,I^2)$.
Then, a sampled $\hat{\theta}\in\mathbb{R}^K$ is derived as:
\begin{equation}
    \hat{\theta} = f(\hat{z}) =  \mathrm{softmax}(W_\theta\hat{z}+b_\theta).
    \label{eq1}
\end{equation}
$W_\theta$, $b_\theta$ are trainable parameters and $K$ denotes the number of topics. Then, we define $\beta\in\mathbb{R}^{K\times |V|},\phi\in\mathbb{R}^{K\times H},e\in\mathbb{R}^{|V|\times H}$ to represent topic-word distributions, topic vectors, and word vectors, respectively. Here, $H$ is the dimension of vectors. $\phi$ is randomly initialized and $e$ can be pre-trained word embeddings. $\beta$ is computed with $\phi$ and $e$ as follows:
\begin{equation}
    \beta_k = \mathrm{softmax}(e\cdot\phi_k^{\top}).
\label{eq2}
\end{equation}
In the generative part, we parameterize $p(d|\beta,\theta)$ and define the loss function of neural topic model as:
\begin{align}
    \mathcal{L}_T & = D_{KL}[q(\theta|d)||p(\theta)]-\mathbb{E}_{q(\theta|d)}[\mathrm{log}p(d|\beta,\theta)] \nonumber \\
    & \approx D_{KL}[q(z|d)||p(z)]-\sum\nolimits_{n}\mathrm{log}p(w_n|\beta,\hat{\theta}).
    \label{eq4}
\end{align}
The first term uses the KL-divergence to ensure that the variational distribution $q(\theta|d)$ is similar to the true prior $p(\theta)$, where $p(\theta)$ represents a standard Gaussian prior $\mathcal{N}(0,I^2)$. In the second term, $w_n$ denotes the $n$-th observed word in $d$ and the log-likelihood of $d$ can be computed with $\mathrm{log}(\hat{\theta}\cdot\beta)$.

{\bf Learning Word-Saliency Correspondences.} In spoken dialogues, abundant noise and common semantics appear randomly and co-occur with informative words. Meanwhile, salient information is encapsulated in dialogue summaries. We therefore assume that each dialogue is a mixture of informative words and other words, where words corresponding to the gold summary are basically informative. We split $K$ topics into two groups: {\em informative topics} and {\em other topics}, where the topic number is $K_s$ and $K_o$ ($K=K_s+K_o$), respectively. Given $\hat{z}$ derived from $d$, the distribution over {\em informative topics} $\hat{\theta}_s\in\mathbb{R}^{K_s}$ and {\em other topics} $\hat{\theta}_o\in\mathbb{R}^{K_o}$ are inferred by $f_s(\cdot)$ and $f_o(\cdot)$ with different parameters:
\begin{align}
    &\hat{\theta}_s = f_s(\hat{z}) = \mathrm{softmax}(W_{\theta_s}\hat{z}+b_{\theta_s}), \nonumber \\
    &\hat{\theta}_o = f_o(\hat{z}) = \mathrm{softmax}(W_{\theta_o}\hat{z}+b_{\theta_o}).
    \label{eq5}
\end{align}
In the generative part of topic modeling, we use $s\in\mathbb{R}^{|V|}$ to represent a word subset of $d$, in which each word appears in both the gold summary and the original dialogue. Thus the parameterization of $p(d|\beta,\theta)$\footnote{Notably, we do not parameterize $p(\theta|s)$ and $p(s|\beta,\theta)$ directly, because gold summaries are not available at test time.} is decomposed by:
\begin{equation}
    p(d|\beta,\theta) = p(s|\beta_s,\theta_s)p(d-s|\beta_o,\theta_o),
    \label{eq6}
\end{equation}
where $\beta_s\in\mathbb{R}^{K_s\times |V|},\beta_o\in\mathbb{R}^{K_o\times |V|}$ along with $\phi_s\in\mathbb{R}^{K_s\times H},\phi_o\in\mathbb{R}^{K_o\times H}$ are constructed by splitting $\beta$ and $\phi$ based on the two topic groups. Eq.\ref{eq6} indicates that topic assignments for summary words are constrained in the group of {\em informative topics}, while other words in the dialogue are gathered by {\em other topics}. As a result, each word in a dialogue is associated with either saliency or not salient. Accordingly, given $w_n^s$ and $w_n^{d-s}$ that denote the $n$-th observed word in $s$ and $d-s$, respectively, the loss function becomes: 
\begin{align}
    \mathcal{L}_T \approx &-\sum\nolimits_{n}\mathrm{log}p(w_n^s|\beta_s,\hat{\theta}_s) -\sum\nolimits_{n}\mathrm{log}p(w_n^{d-s}|\beta_o,\hat{\theta}_o) \nonumber \\
    & + D_{KL}[q(z|d)||p(z)].
    \label{eq7}
\end{align}
Compared to NTM, the proposed SATM leverages dialogue summaries to detach informative words from noise and common semantics, avoiding the direct topic modeling on a mixture of useful and useless content. Hence, noise and common semantics can hardly obscure the underlying informative words, making the topic inference more robust. Besides, SATM can be easily employed as an external module and combined with other summarization models like Wang et al. \shortcite{wang2018reinforced} as long as gold summaries are available.

{\bf Multi-Role Topic modeling.} 
Based on SATM, we input $d$ and infer topic representations $t_s\in\mathbb{R}^H$ and $t_o\in\mathbb{R}^H$ by:
\begin{equation}
    t_s = \phi_s^{\top}\cdot\hat{\theta}_s, \quad\quad t_o = \phi_o^{\top}\cdot\hat{\theta}_o.
    \label{eq8}
\end{equation}
Here, $t_s$ can be regarded as a topic vector that captures informative topic information while $t_o$ gathers noise and common semantics, both of which will be incorporated into the TDS to facilitate the saliency estimation. Furthermore, in order to capture role-specific information, we perform topic modeling on customer utterances and agent utterances separately. Given the bag-of-words representation of customer utterances $d^C\in\mathbb{R}^{|V|}$ and agent utterances $d^A\in\mathbb{R}^{|V|}$, we can infer topic distributions $\hat{\theta}^C_s,\hat{\theta}^C_o,\hat{\theta}^A_s,\hat{\theta}^A_o$ using Eq.\ref{eq5}. Then, topic representations for different roles $t^C_s,t^C_o,t^A_s,t^A_o$ can be obtained similar to Eq.\ref{eq8}. Hence, we have totally three topic models with different parameters on customer utterances, agent utterances, and overall dialogues, respectively.

\subsection{Topic-Augmented Two-Stage Dialogue Summarizer}

In the customer service scenario, spoken dialogues are long and sometimes twisted, where most utterances are unimportant or even noisy, which can be directly filtered out. Hence, we follow Chen and Bansal \shortcite{chen2018fast} and employ a two-stage summarizer that first selects salient utterances and then refines them. The basic two-stage summarizer consists of an {\em utterance extractor} and an {\em abstractive refiner}. The extractor encodes each utterance $u_i$ into an utterance representation $h_i$ and employs a {\em Pointer Network} \cite{vinyals2015pointer} to recurrently extract utterances based on $h_i$. The refiner is a standard sequence-to-sequence (Seq2seq) model, which can generate a concise summary based on the extracted utterances. To bridge the extractor and the refiner, a policy gradient technique \cite{williams1992simple} is applied to train the overall summarizer, where we use $\mathcal{L}_{S}$ to represent the loss function. 

In this work, we use Transformer \cite{vaswani2017attention} as the basic encoder and decoder layer for TDS. For the $i$-th utterance in a dialogue, we have $u_i=\{r_i,e_{i1},..,e_{iN}\}$, where $r_i$ is a role embedding representing the speaker of $u_i$, which can be either {\em customer} or {\em agent}. $e_{ij}$ is the embedding of the $j$-th word. For more details of the basic two-stage model and our implementation, please refer to Chen and Bansal \shortcite{chen2018fast} and the supplementary due to the space limitation.

{\bf Topic Information Augmentation.} To capture role information and highlight global topics, we incorporate multi-role topic representations into the Transformer Decoder via a topic-informed attention mechanism for saliency estimation, which is an extension of multi-head attention \cite{vaswani2017attention}. Formally, let $q_i$ denote the $i$-th decoding step of the query, and $x_j$ denote the $j$-th element in the memory, the original attention mechanism for each head is defined as:
\begin{align}
    \alpha^q_{ij} &= \mathrm{softmax}((q_iW_Q)(x_jW^q_{K})^\top/\sqrt{d_{h}}), \nonumber \\
    \mu^q_i &= \sum\nolimits_j\alpha^q_{ij}(x_jW_V),
\end{align}
where $W_Q,W^q_{K},W_V$ are trainable parameters and $d_h$ is the dimension of each head. $\mu^q_i$ is a vector that fuses salient information based on the query at the $i$-th decoding step. In a general decoding process, the state of fused memory at each step is conditioned on the previously decoded sequence where errors may be accumulated. Here, we additionally use global topics and role information as a guidance to assist with sequence decoding by measuring relevance between memory elements and multi-role topics. Formally, we design an auxiliary attention operation as follows:
\begin{align}
    \alpha^t_{j} &= \mathrm{softmax}((\tau_sW_T-\tau_oW_T)(x_jW^t_{K})^\top/\sqrt{d_h}), \nonumber \\
    \mu^t &= \sum\nolimits_j\alpha^t_{j}(x_jW_V).
    \label{eq10}
\end{align}
Here, $\tau_s$ and $\tau_o$ represent role-specific topic representation, where $\tau_s=[t_s;t_s^C;{\bf0}],\tau_o=[t_o;t_o^C;{\bf0}]$ if $x_j$ corresponds to the {\em customer} speaker, and $\tau_s=[t_s;{\bf0};t^A_s],\tau_o=[t_o;{\bf0};t^A_o]$ if $x_j$ corresponds to the {\em agent} speaker. $[\cdot;\cdot]$ means concatenation and ${\bf0}$ is a vector with all elements set to 0. In Eq.\ref{eq10}, we design $\alpha^t_j$ that makes $\tau_s$ contrary to $\tau_o$ inspired by the contrastive attention \cite{duan2019contrastive}, which encourages the attention to topic-relevant elements, and discourages the attention to noise and common semantics. Hence, $\tau_s$ and $\tau_o$ work in an opposite way to contribute to an overall target. $\mu^t$ can be regarded as a topic-aware vector that basically discards noisy and uninformative elements in the memory. Finally, we combine the above two attention operations to consider both the global topics and the current query at each decoding step, and form an integrated memory fusion $\mu_i$ by:
\begin{align}
    p^{sel}_i &= \sigma([q_i;\mu^q_i;\mu^t]\cdot W_P), \nonumber \\
    \alpha_{ij} &= (1 - p^{sel}_i) \cdot \alpha^q_{ij} + p^{sel}_i \cdot \alpha^t_j, \nonumber \\
    \mu_i &= \sum\nolimits_j\alpha_{ij}(x_jW_V),
\end{align}
where $p^{sel}_i\in(0,1)$ denotes the selective probability used as a soft switch to choose between the original query-based attention or the topic-guided attention. The attention mechanism is further adapted into the multi-head manner similar to Vaswani et al. \shortcite{vaswani2017attention}. Notably, we apply the topic-informed attention mechanism to both the utterance extractor and the abstractive refiner. For the extractor, $x_j$ represents the hidden state of $j$-th utterance. For the refiner, $x_j$ is the hidden state of $j$-th word in selected utterances. 
As a result, we can perform saliency estimation assisted by multi-role topic information on both the utterance level and word level.

\subsection{Joint Training} 
To jointly train the summarizer and the multi-role topic models, we design a joint loss which includes the loss function of summarizer $\mathcal{L}_{S}$ and the loss functions of three topic models $\mathcal{L}^C_T,\mathcal{L}^A_T,\mathcal{L}_T$ that correspond to customer utterances, agent utterances and overall dialogues, respectively. The joint loss function is defined as:
\begin{equation}
    \mathcal{L} = \mathcal{L}_{S} + \lambda(\mathcal{L}^C_T+ \mathcal{L}^A_T + \mathcal{L}_T),
\end{equation}
where $\lambda$ is a coefficient to balance the losses between the summarizer and topic models.

\section{Experimentation}
In this section, We describe the experiments conducted on a real-world customer service dataset. We compare our proposed TDS+SATM with strong baselines and further analyze the influence of different parts in our model. All hyper-parameters tuning is conducted on the validation set. Full training details can be found in the supplementary.
\begin{table}[t!]
\footnotesize
\begin{center}
\setlength{\tabcolsep}{2mm}{
\begin{tabular}{c|cc|ccc}
\toprule[1pt]
 &\bf \# of& \bf \# of & \multicolumn{3}{c}{\bf average token number}\\
 & \bf dialogues & \bf utterances & \em dialog. & \em utter. & \em summ.\\
\midrule
Train & 17,189 & 872,292 & 1,333.72  & 26.28& 54.54\\
Dev. & 820 & 38,461 & 1,221.12  & 26.03 & 53.73   \\
Test & 851 & 42,667 & 1,300.98 & 25.95 & 54.42 \\
\bottomrule[1pt]
\end{tabular}}
\end{center}
\caption{\label{data}Statistics of the customer service dataset. }
\end{table}

\subsection{Dataset} 
Our customer service dialogue dataset is collected from the call center of an E-commerce company. All dialogues are incoming calls in Mandarin Chinese that take place between a customer and a service agent. After each round of service, an agent needs to write a brief description about the conversation, which mainly includes problems the customer faces and solutions the agent provides. For each dialogue example, we take the agent-written description as the gold summary. All dialogues are originally in the form of audio and we transcribe them into texts using an ASR model pre-trained on customer service dialogues \cite{zhang2019investigation} with a character error rate of 9.3\%. The final dataset therefore includes dialogue-summary pairs that consist of dialogue transcriptions and human-written summaries. We totally collect 18.86K dialogues with 953K utterances and split them into training (90\%), development (5\%), and test (5\%) set. Table \ref{data} shows the detailed statistics of the collected dataset. 

\begin{table}[t!]
\small
\begin{center}
\begin{tabular}{lcccc}
\toprule[1pt]
\bf Methods &\bf RG-1 & \bf RG-2 & \bf RG-L & \bf BLEU \\
\midrule[1pt]
Ext-Oracle & 41.38 & 15.64 & 29.18 & 7.28\\
\midrule
Seq2seq+Att & 28.66 & 13.05 & 22.63 & 6.89\\
PGNet & 34.88 & 17.81& 27.80& 9.77\\
TRF & 35.17& 18.01& 28.05& 9.87\\
CopyTRF & 34.97& 17.84 & 27.88& 9.78\\
HiBERT* & 35.50 & 18.24 &28.44 & 9.89\\
BERT+TRF* & 35.67 & 18.49 & 28.57& 10.19\\
FastRL* & 35.99 & 18.67 &28.86 & 10.40\\
\cellcolor{mygray}TDS+NTM (base)& \cellcolor{mygray}35.21 & \cellcolor{mygray}18.04& \cellcolor{mygray}28.11& \cellcolor{mygray}9.87\\
\cellcolor{mygray}TDS+SATM (base)& \cellcolor{mygray}35.75 & \cellcolor{mygray}18.54& \cellcolor{mygray}28.62& \cellcolor{mygray}10.34\\
\cellcolor{mygray}TDS+NTM*& \cellcolor{mygray}36.13 & \cellcolor{mygray}19.09& \cellcolor{mygray}29.00& \cellcolor{mygray}10.77\\
\cellcolor{mygray}TDS+SATM* & \cellcolor{mygray}\bf 36.81 &\cellcolor{mygray}\bf 19.63& \cellcolor{mygray}\bf 29.61& \cellcolor{mygray}\bf 11.24 \\
\bottomrule[1pt]
\end{tabular}
\end{center}
\caption{\label{main_results} Results of automatic metrics on the customer service dataset. RG-(1,2,L) represents the F1 score of ROUGE-(1,2,L). TRF denotes the Transformer. Methods marked with * utilize BERT as the word-level encoder.}
\end{table}
\subsection{Comparison Methods} 
\begin{itemize}
    \item {\bf Ext-Oracle \cite{nallapati2017summarunner}}, where a greedy algorithm is applied to select utterances whose combination maximizes the evaluation score against the gold summary, which is used as the upper bound of extractive methods.
    \item {\bf Seq2seq+Att \cite{nallapati2016abstractive}} is a standard RNN-based encoder-decoder model with attention mechanisms.
    \item {\bf PGNet \cite{see2017get}} has a pointer mechanism where the decoder can choose to generate a word from the vocabulary or copy a word from the source text.
    \item {\bf Transformer \cite{vaswani2017attention}} is an attention-based model, for which we also implement a variant with the copy mechanism, denoted as CopyTransformer. 
    \item {\bf BERT+Transformer \cite{liu2019text}} consists of a BERT encoder \cite{devlin2019bert}\footnote{The original BERT is only applicable for texts with a maximum length of 512. We extend the range of positional embeddings to make it possible to encode long dialogues.}
    and a Transformer decoder. They are tuned with different optimizers to alleviate the mismatch between BERT and other parameters\footnote{Other models equipped with BERT use the same strategy.}.
    \item {\bf HiBERT \cite{zhang2019hibert}} can encode documents on the word level and sentence level hierarchically. Here, we replace the word-level encoder with BERT and add a basic Transformer decoder to enable Seq2seq learning.
    \item {\bf FastRL \cite{chen2018fast}} is the basic two-stage framework. We implement it based on Transformers and pre-train it with BERT, in that Transformer encoder can be easily combined with pre-trained LMs. 
    \item {\bf TDS+Topic Model} is our approach with different topic models, including NTM and SATM. For a fair comparison, we also employ BERT as the word-level encoder.
\end{itemize}

\subsection{Automatic Evaluation} 
Table \ref{main_results} shows the automatic evaluation results on the customer service dataset (examples of system output can be found in the supplementary). We evaluate summarization quality using ROUGE F1 \cite{lin2004rouge} and BLEU \cite{papineni2002bleu}. We use unigram and bigram overlap (ROUGE-1, ROUGE-2) between system outputs and gold summaries to assess informativeness, and the longest common subsequence (ROUGE-L) to assess fluency. For BLEU, we use 4-grams at most and average the scores of different grams. All metrics are computed on Chinese characters to avoid the influence of word segmentation. 
\begin{table}[t!]
\small
\begin{center}
\begin{tabular}{lcc}
\toprule[1pt]
\bf Methods &\bf Informativeness & \bf Fluency \\
\midrule[1pt]
PGNet & -0.196 & -0.248\\
BERT+TRF & -0.116 & -0.042 \\
HiBERT & -0.120 & -0.030 \\
FastRL & -0.084 & \ 0.086 \\
TDS+SATM & \ \bf0.032 & \ \bf0.098\\
\midrule
Gold & \ 0.484 & \ 0.136 \\
\bottomrule[1pt]
\end{tabular}
\end{center}
\caption{\label{human} Human evaluation with system ranking results.}
\end{table}

The first block in the table includes Oracle as an upper bound for extractive methods. The second block shows the results of abstractive models. From Table \ref{main_results} we can see that most abstractive methods achieve competitive results or even outperform Ext-Oracle on ROUGE-(2/L) and BLEU. It provides evidence that our dataset collects highly abstractive summaries, which requires a system to integrate dialogue content and produce a coherent discourse. The basic TDS plus topic models achieves competitive results against other baselines and outperforms some BERT-based models when equipped with SATM. After combining BERT, two-stage systems (FastRL / TDS) show superior performances compared to other Seq2seq approaches, which probes the effectiveness of extract-refine strategy and indicates that useful information is diluted in long spoken dialogues. When topic information is incorporated, results are further improved. TDS+NTM uses the basic neural topic model and removes the contrastive mechanism in Eq.\ref{eq10}, which brings slight improvements against FastRL. In contrast, TDS+SATM leads to a significant performance enhancement over FastRL on ROUGE (+0.82, +0.96, +0.75 on ROUGE-1/2/L) and BLEU (+0.84) with $p<$ 0.05. It validates that topic information is beneficial for summarizing customer service dialogues, and SATM with word-saliency learning can further boost overall results by improving the quality of topic modeling.

\subsection{Human Evaluation} Metrics for automatic evaluation based on n-grams may not truly reflect the quality of generated summaries. Hence, we further randomly sample 100 examples in the test set for human evaluation. We follow Narayan at al. \shortcite{narayan2018don} to design the experiment, in which three volunteers are invited to compare summaries produced from PGNet, BERT+TRF, HiBERT, FastRL, our proposed TDS+SATM, and the gold summary (Gold). Each volunteer is presented with a dialogue and two summaries produced from two out of six systems and is asked to decide which summary is better in order of two dimensions: {\bf informativeness} (which summary captures more important information in the dialogue?) and {\bf fluency} (which summary is more fluent and well-formed?). We collect judgments from three volunteers for each comparison with the order of dialogues and summaries randomized.

\begin{table}[t]
\small
\begin{center}
\begin{tabular}{lcccc}
\toprule[1pt]
\bf Methods &\bf RG-1 & \bf RG-2 & \bf RG-L & \bf BLEU \\
\midrule[1pt]
TDS+SATM & 36.81 &\bf 19.63 &\bf 29.61 &\bf 11.24\\
\ \ (w/o) Cust. &\bf 36.84&19.60 &29.56 &11.09 \\
\ \ (w/o) Agent&36.79 & 19.39& 29.50& 10.73\\
\ \ (w/o) Agent \& Cust. & 36.37& 19.03 & 29.10& 10.37\\
\bottomrule[1pt]
\end{tabular}
\end{center}
\caption{\label{ablation} Ablation study of TDS+SATM with different kinds of topic modeling. Agent and Cust. represent topic modeling on agent utterances and customer utterances, respectively.}
\end{table}
\begin{figure}
\centering
  \includegraphics[width=3.0in]{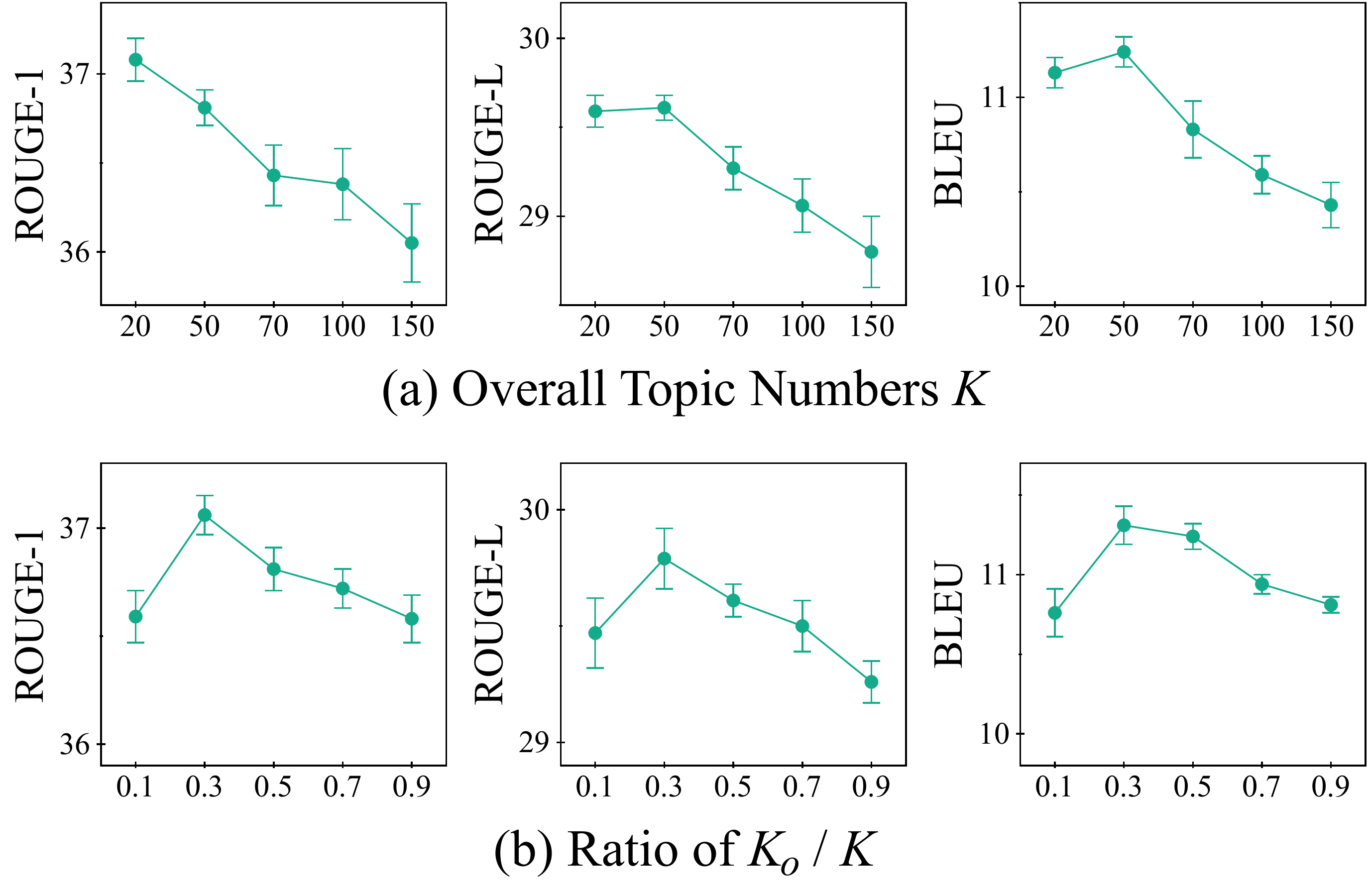}
  \caption{Effects of different topic numbers of SATM.} \label{fig:topic_num}
\end{figure}
Table \ref{human} gives the system ranking results of human evaluation. The score of a system is calculated as the percentage of times it was selected as best minus the percentage of times it was chosen as worst, which ranges from -1 (worst) to 1 (best). Gold summaries are unsurprisingly ranked best on both two dimensions. For informativeness, TDS+SATM ranks second followed by other systems, which validates the effectiveness of our proposed SATM and topic argumentation mechanism for assisting with saliency estimation. In terms of fluency, two-stage models (FastRL / TDS) are considered better than other baselines, which indicates that the extract-refine strategy can generate more fluent summaries. We also conducted pairwise comparisons between systems (using a binomial two-tailed test; null hypothesis: two systems are equally good; $p<$0.05). In terms of informativeness, TDS+SATM is significantly different from all other systems. In terms of fluency, two-stage systems are significantly different from other systems, and FastRL is not significantly different from TDS+SATM. 

\subsection{Analysis and Discussion}
To better understand the influence of role information, saliency-aware topic modeling, and the topic-informed attention, we perform the following qualitative analysis.

{\bf Contribution of Role Information.} Table \ref{ablation} shows the results of TDS+SATM with different kinds of topic modeling. When we remove one of the topic models on customer utterances or agent utterances, results are not be appreciably affected. However, after removing both two topic models, the system suffers a significant performance degradation ($p<$0.05). It indicates that role-specific information is beneficial for dialogue modeling, and at least one of the speakers should be specified to make role content distinguishable. 

{\bf Effect of Topic Numbers in SATM.} The topic number is a critical hyper-parameter in topic models because it potentially affects the convergence rate and the inference quality. Here, we report results of TDS+SATM with different topic numbers in Figure \ref{fig:topic_num}. Figure \ref{fig:topic_num}(a) shows the effects of $K$ that ranges from 20 to 150 with $K_s=K_o$. It shows a performance decline trend and a variance increase trend when $K$ is continuously increased after exceeding 50. It indicates that a proper topic number is sufficient for capturing main topics and a larger one makes the topic inference unstable. Figure \ref{fig:topic_num}(b) shows the effects of different ratios of $K_o/K$, where we fix $K$=50 and adjust $K_o$ in range of 5 to 45. The results show that an overly unbalanced number of $K_s$ and $K_o$ can hurt the performance. It indicates that each dialogue is a mixture of useful and uninformative content, either of which can not be ignored when performing topic modeling on dialogues.

\begin{table}[t]
\small
\begin{center}
\begin{tabular}{|c|l|}
\hline
\multirow{6}{*}{\tabincell{c}{\tabincell{c}{SATM \\ Informative \\ Topic}}} & {\bf T1:} deliver, time, order, address, modify,\\
& \quad\ \ \  cancel, ship, return, refund, receive\\
& {\bf T2:} feedback, problem, submit, suggest, apply, \\
& \quad\ \ \  complain, seller, quality, product, slow\\
& {\bf T3:} buy, account, pay, bind, phone, number,\\
& \quad\ \ \  modify, check, username, message\\
\hline
\multirow{4}{*}{\tabincell{c}{\tabincell{c}{SATM \\ Other \\ Topic}}} & {\bf T1:} please, wait, service, sorry, really, thanks, \\
& \quad\ \ \ bother, mean, find, welcome\\
& {\bf T2:} send, call, record, again, later, check, help,\\
& \quad\ \ \ keep, contact, reply\\
\hline
\multirow{4}{*}{\tabincell{c}{\tabincell{c}{NTM \\ General \\ Topic}}}& {\bf T1:} thanks, later, sorry, please, really, phone,\\
& \quad\ \ \ feedback, deliver, number, return\\
& {\bf T2:} apply, sorry, again, order, check, wait, \\
& \quad\ \ \ record, reply, seller, contact\\
\hline
\end{tabular}
\end{center}
\caption{\label{topic_words} Top-10 words of example topics in different topic groups learned by joint training of TDS+Topic Model.}
\end{table}

\begin{figure}[t]
\centering
  \includegraphics[width=2.4in]{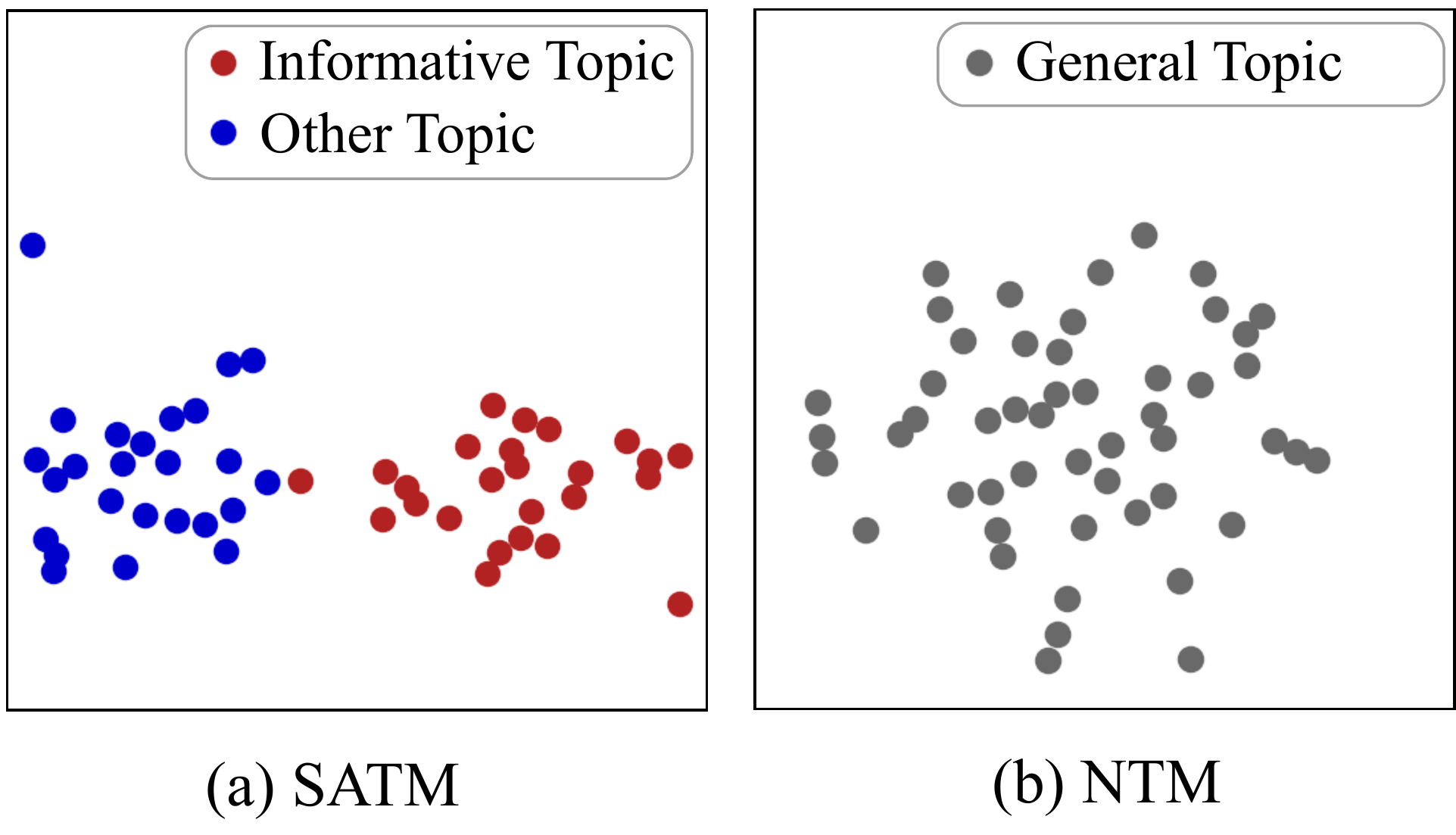}
  \caption{2-D t-SNE visualizations of topic vectors ($K$ = 50 and $K_s$ = $K_o$ = 25).} \label{fig:topic_vec}
\end{figure}

{\bf Comparison between NTM and SATM.} To thoroughly compare the standard NTM and our proposed SATM, we analyze the topic-word distributions $\beta$ and the topic vectors $\phi$ learned by TDS+Topic Model. Table \ref{topic_words} shows topic examples of different topic groups, where top-10 words with the highest probability in $\beta$ are listed\footnote{They are translated from Chinese with stop words removed.}. We found that words in {\em informative topics} can better reflect specific dialogue scenes. For instances, topic 1 with {\em address}, {\em ship} and {\em refund} is about delivery issues. Topic 3 with {\em account}, {\em bind} and {\em username} is about account issues. By contrast, {\em other topics} tend to gather noise and common semantics, where words of topic 1 often appear in unrelated chit-chats, and topic 2 includes common words in the customer service scenario. As for {\em general topics} in NTM, top-10 words show a mixture of informative and common content. Besides, we analyze the topic vectors $\phi$ and visualize the latent space in 2-D using t-SNE \cite{maaten2008visualizing} for SATM and NTM. In Figure \ref{fig:topic_vec}(a), with the learning of word-saliency correspondences and the contrastive mechanism, vectors of two topic groups in SATM are effectively mapped to separate regions, while in Figure \ref{fig:topic_vec}(b), topic vectors of NTM do not show obvious clusters.

\begin{figure}[t]
\centering
  \includegraphics[width=3.3in]{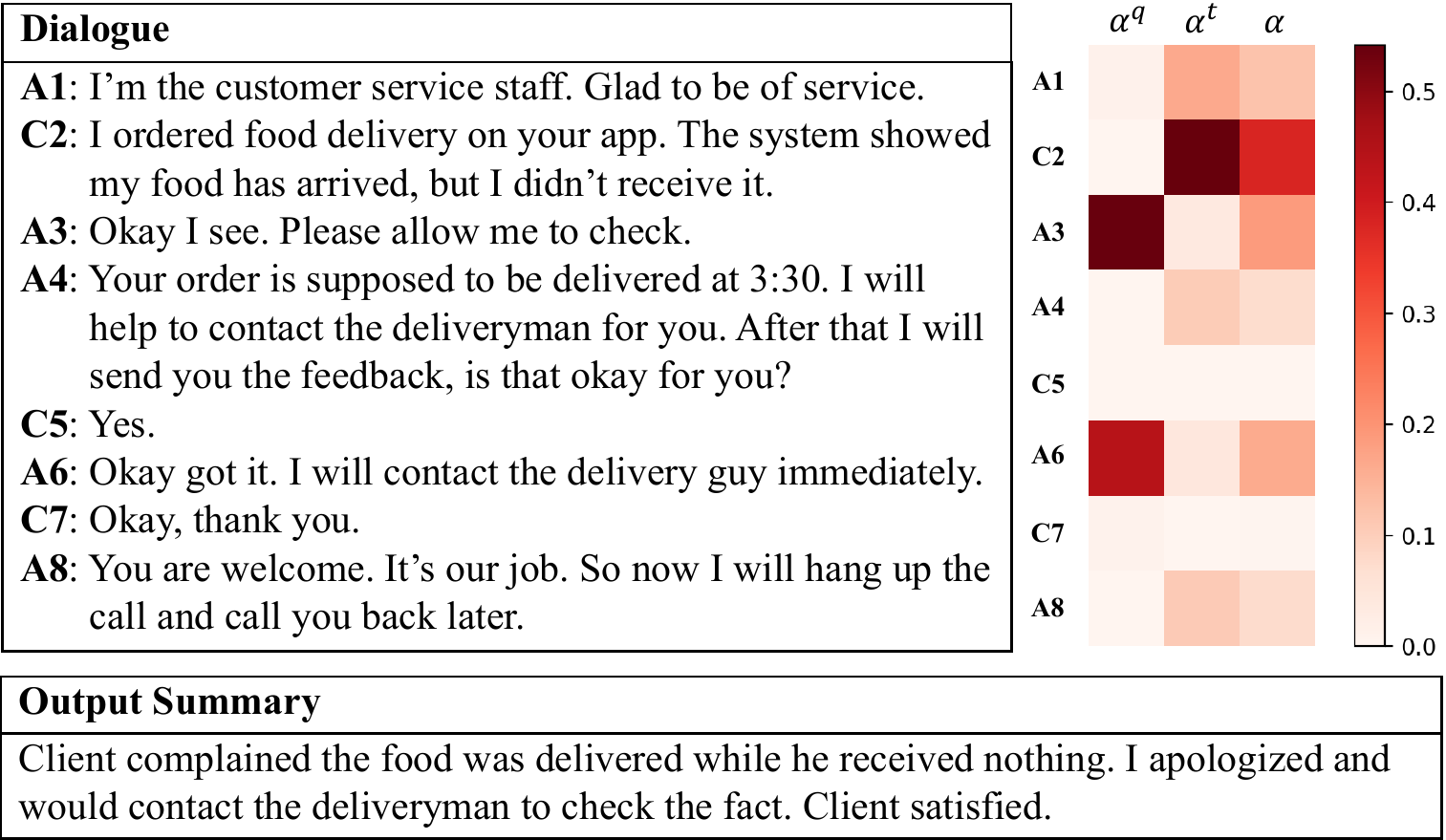}
  \caption{Utterance-level attention map of an example dialogue along with the output summary from TDS+SATM.} \label{fig:attention}
\end{figure}

{\bf Case Study of Topic-Informed Attention Mechanism.} Figure \ref{fig:attention} shows the attention map of an examplar dialogue with a translated summary generated by TDS+SATM. $\alpha^q$, $\alpha^t$, $\alpha$ represent the query-based attention, the topic-guided attention and the final combined attention, respectively. Attention scores are taken from the decoder of extractor to demonstrate the beginning step in an utterance-level decoding process. From the example we can see that topic-guided attention $\alpha^t$ successfully focuses on the salient utterance that mentions customer's problem. Then the combined attention $\alpha$ exhibits preference to $\alpha^t$ and focuses on appropriate utterances that finally contribute to the summary generation.

\section{Related Work}
\subsection{Dialogue Summarization}
 
Dialogue summarization is a challenging task and has been widely explored in various scenarios. Previous works generally focus on summarizing dialogues by stringing key points to maintain an integral dialogue flow: Mehdad et al. \shortcite{mehdad2013abstractive} and Shang et al. \shortcite{shang2018unsupervised} 
first group utterances that share similar semantics by community detection, and then generate a summary sentence for each utterance group. Liu et al. \shortcite{liu2019automatic} propose a hierarchical model to produce key point sequences and generate summaries at the same time for customer service dialogues. Duan et al. \shortcite{duan2019legal} train the assignment of utterances to the corresponding controversy focuses to summarize court debate dialogues. Several works \cite{zechner2001automatic,xie2008evaluating,oya2014template,liu2019topic,li2019keep} split dialogues into multiple segments by means of topic segmentation when conducting summarization. Different from above works, Pan et al. \shortcite{pan2018dial2desc} first attempt to generate a highly abstractive summary for the entire dialogue, which produces a concise event/object description with a Transformer-based approach. By contrast, in this work, dialogue summaries generally highlight role-specific content, which requires the system to further focus on the role information when performing saliency estimation.

\subsection{Text Summarization with Topic Modeling}
Topic models have been extensively studied for document modeling and information retrieval. Probabilistic topic models like pLSA \cite{hofmann1999probabilistic} and LDA \cite{blei2003latent} provide a theoretically sound foundation for uncovering the underlying semantics of a document. Recently, neural topic models \cite{miao2017discovering} have been introduced to infer latent representations for documents, which leverage deep neural networks as approximators for learning topic distributions. A couple of works have employed these topic models to facilitate the summarization task. Wang et al. \shortcite{wang2018reinforced} and Narayan et al. \shortcite{narayan2018don} use LDA to infer topic embeddings and design a joint attention mechanism to incorporate topic information. Fu et al. \shortcite{fu2020document} merge the topic inference module with a summarization model rather than simply resort to using a pre-trained topic model. Some early works on dialogue summarization \cite{higashinaka2010learning,wang2012unsupervised,sood2013topic} directly perform topic modeling on dialogues to extract salient words or utterances. All the above methods leverage the standard topic modeling framework as an auxiliary tool to conduct topic mining for summarization. 
By contrast, we use summary information as a guidance to force the topic model to learn word-saliency correspondences. As a result, underlying semantics can hardly be obscured by uninformative content, making the salient information more perceivable.

\section{Conclusion and Future Work}
In this paper, we propose a topic-augmented two-stage summarizer with a multi-role topic modeling mechanism for customer service dialogues, which can generate highly abstractive summaries that highlight role-specific information. Moreover, we introduce a novel training regime for topic modeling that directly learns word-saliency correspondences to alleviate the influence of uninformative content. Experiments on a real-world customer service dataset validate the effectiveness of our approach. Future directions may be the exploration of template-guided abstractive methods to make summaries more standardized and easier for reporting. 

\section{Acknowledgments}
The authors wish to thank the anonymous reviewers for their helpful comments. This work was partially funded by China National Key R\&D Program (No. 2018YFC0831105), National Natural Science Foundation of China (No. 61751201, 62076069, 61976056), Shanghai Municipal Science and Technology Major Project (No.2018SHZDZX01), Science and Technology Commission of Shanghai Municipality Grant  (No.18DZ1201000, 17JC1420200). This work was supported by Alibaba Group through Alibaba Innovative Research Program. 
\bibstyle{aaai}
\bibliography{main}
\end{document}